\documentclass{article}

\usepackage[nonatbib,preprint]{neurips_2024}

\usepackage{amsmath}
\usepackage{macros}
\usepackage{graphicx}
\usepackage{caption}
\usepackage{subcaption}
\usepackage{algorithm2e}

\usepackage[utf8]{inputenc} 
\usepackage[T1]{fontenc}    
\usepackage{hyperref}       
\usepackage{url}            
\usepackage{booktabs}       
\usepackage{amsfonts}       
\usepackage{nicefrac}       
\usepackage{microtype}      
\usepackage{xcolor}         

\newcommand{\ars}[1]{\renewcommand{\arraystretch}{#1}}

\title{A logical alarm for misaligned binary classifiers}

\author{%
  Andr\'es Corrada-Emmanuel \\
  3D Rationality \\
   \And
  Ilya Parker \\
  3D Rationality \\
  \AND
  Ramesh Bharadwaj \\
  U.S.\ Naval Research Laboratory \\
}

\begin{document}

\maketitle

\begin{abstract}
  If two agents disagree in their decisions, we may suspect they
  are not both correct. This intuition is formalized for evaluating
  agents that have carried out a binary classification task. Their
  agreements and disagreements on a joint test allow us to
  establish the only group evaluations logically consistent with their
  responses. This is done by establishing a set of axioms (algebraic
  relations) that must be universally obeyed by all evaluations of
  binary responders. A complete set of such axioms are possible for
  each ensemble of size N. The axioms for $N = 1, 2$ are used to
  construct a fully logical alarm - one that can prove that at least
  one ensemble member is malfunctioning using only unlabeled data.
  The similarities of this approach to formal software verification
  and its utility for recent agendas of safe guaranteed AI are discussed.
\end{abstract}

\section{Introduction}

Formal verification of AI systems has recently been proposed as a way to
make them safer\cite{Szegedy2020, Dalrymple2024}. So far, 
these proposals have focused on aspects of machine
training and decision making - how do we train and/or certify AI agents
to make them safer? Here we consider formal verification of unsupervised
agent evaluations, whether human or robotic.

Consider an ensemble of N agents given a task. No matter how complex the task,
we can engage other agents to evaluate or supervise them. This has become a
popular methodology for making safer and more trustworthy LLM systems. 
Weak-to-strong supervision \cite{Burns2023}
has been proposed to tackle the fundamental challenge of aligning 
superhuman models. LLMs criticizing LLM code generators reduce bugs \cite{McAleese2024}.
Adversarial AI debates help weaker or non-expert humans answer questions more accurately
\cite{Irving2018,Khan2024}.
All such
schemes inevitably raise the specter of infinite regression (supervisors that
supervise supervisors that ...) or are unverifiable themselves. This problem
is not inherently an AI problem but rather a classic problem in epistemology
and economics - the principal/agent monitoring problem. Agents whether human
or robotic are employed to carry out tasks. The principal, the one responsible
for giving them the task, does not have the ability or time to supervise them.
How can the principal make sure that the agents are doing their work correctly
and safely?

The approach taken here is that we can formalize unsupervised evaluations so as
to ameliorate this bottleneck problem in operating safe AI systems. Formal
verification of software systems is well known and has notable, direct applications
to the safety of complex engineering systems such as nuclear plants \cite{Lawford2012}.
Here we consider
how to formalize verification of unsupervised evaluations. In such settings
there is no answer key that can help us grade or evaluate noisy agents that have
taken a test. A logic in such settings cannot prove the soundness of group
evaluations. But it can prove their logical consistency - what are the group
evaluations that are consistent with how they responded on the test?

Formal software verification frameworks have three aspects. Dalrymple et al 
\cite{Dalrymple2024} call them - 
the \emph{world model}, the \emph{safety specification}, and the \emph{verifier}.
All these aspects will be discussed using a set of complete polynomials
that generate observed statistics of how classifiers agree and disagree on a test
given how correct they were on it. There are $2^N$ ways that $N$ binary
classifiers could vote on the true label of an item, and for each we can write
a polynomial.

The paper is organized as follows. In part 2 we discuss the polynomial generating
set for the trivial ensemble, one single classifier ($N=1$), and pair
ensembles, ($N = 2$). From the generating set for $N=1$ we will derive a single
``axiom'' or universally true algebraic relation that all members of an ensemble
must satisfy. When we analyze the generating set for a pair of classifiers, we will
find that it contains the single classifier axioms and a single new axiom for the pair.
In part 3 we use the single classifier axiom discussed in part 2 to
create a logical alarm for misaligned binary classifiers. We conclude with a
brief discussion of the use of this formalism when dealing with super-intelligent
agents.

\section{A verification formalism for binary evaluations}

There are many \emph{evaluation models} for a binary response test.
The ones used here are associated with the $2^N$ possible decision
patterns when we observe the joint decisions of an ensemble on a
given item or question. We will
detail the models for the $N=1$ and $N=2$ cases.

\subsection{An evaluation model for the trivial ensemble, $N=1$}

\begin{figure}
  \centering
  \includegraphics[width=0.5\textwidth]{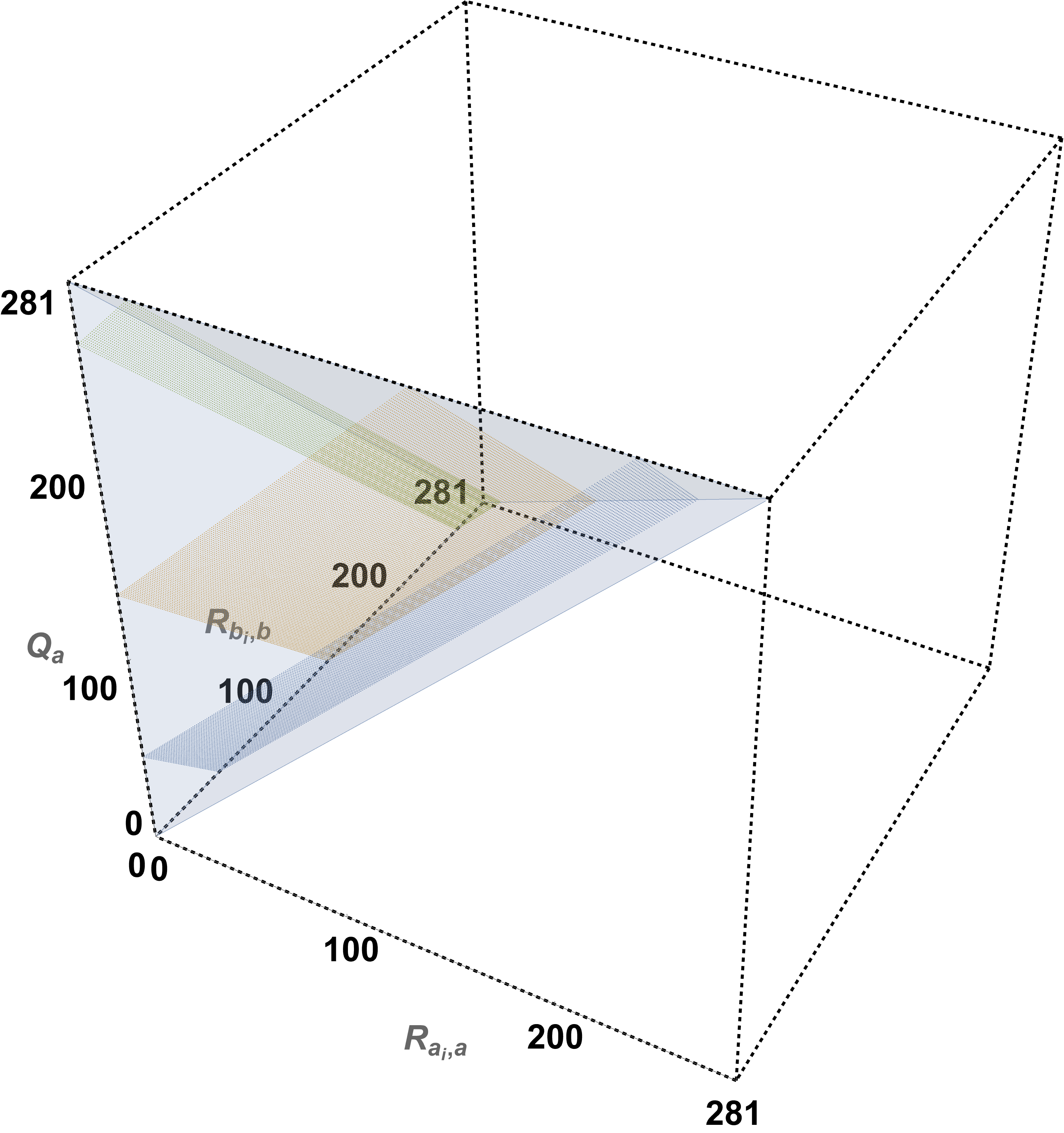}
  \caption{The set of all possible evaluations in (\raia{i},\rbib{i}, \qa) space
  for a $Q=281$ test for three LLMs grading a fourth one on the multistep-arithmetic
  task in the BIG-Bench-Mistake Dataset. Once we observe how the test was answered,
  we can use the single classifier axiom: $(Q \raia{i} - \qa \rai{i}) - (Q \rbib{i} - \qb \rbi{i}).$
  It defines a much smaller set of evaluations consistent with the observed test
  responses. Somewhere in this set is the unknown ground truth value for the
  number of correct responses in each label, \raia{i} and \rbib{i}. Once we know
  the number of responses for a classifier, \rai{i} and \rbi{i}, the axiom defines a
  plane as pictured here for the three grading LLMs.}
  \label{fig:evaluation-cube}
\end{figure}

\begin{figure}
     \centering
     \begin{subfigure}[b]{0.45\textwidth}
         \centering
         \includegraphics[width=\textwidth]{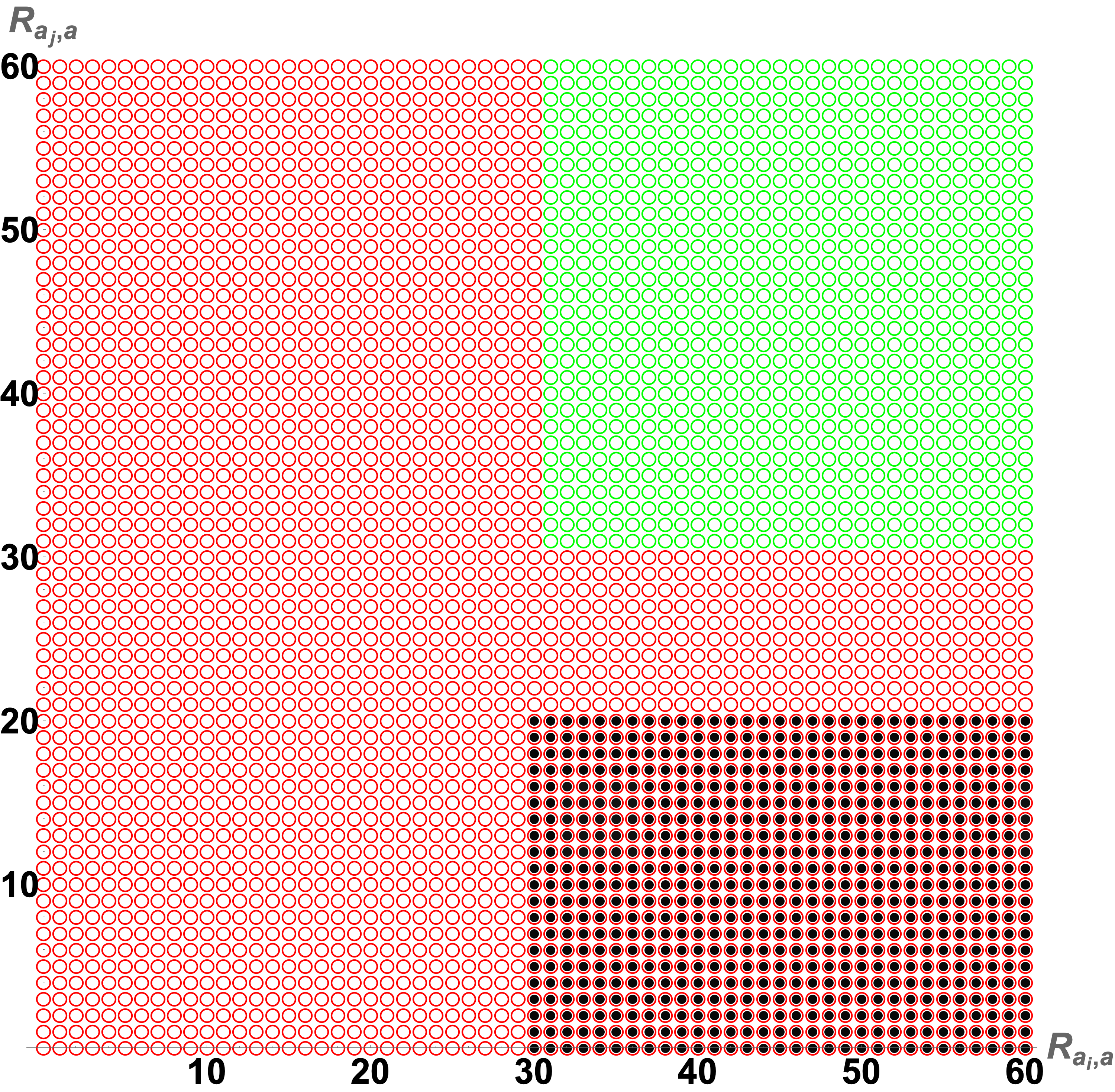}
         \label{fig:aLabelSquare}
     \end{subfigure}%
     \hspace{2pt}
     \begin{subfigure}[b]{0.45\textwidth}
         \centering
         \includegraphics[width=\textwidth]{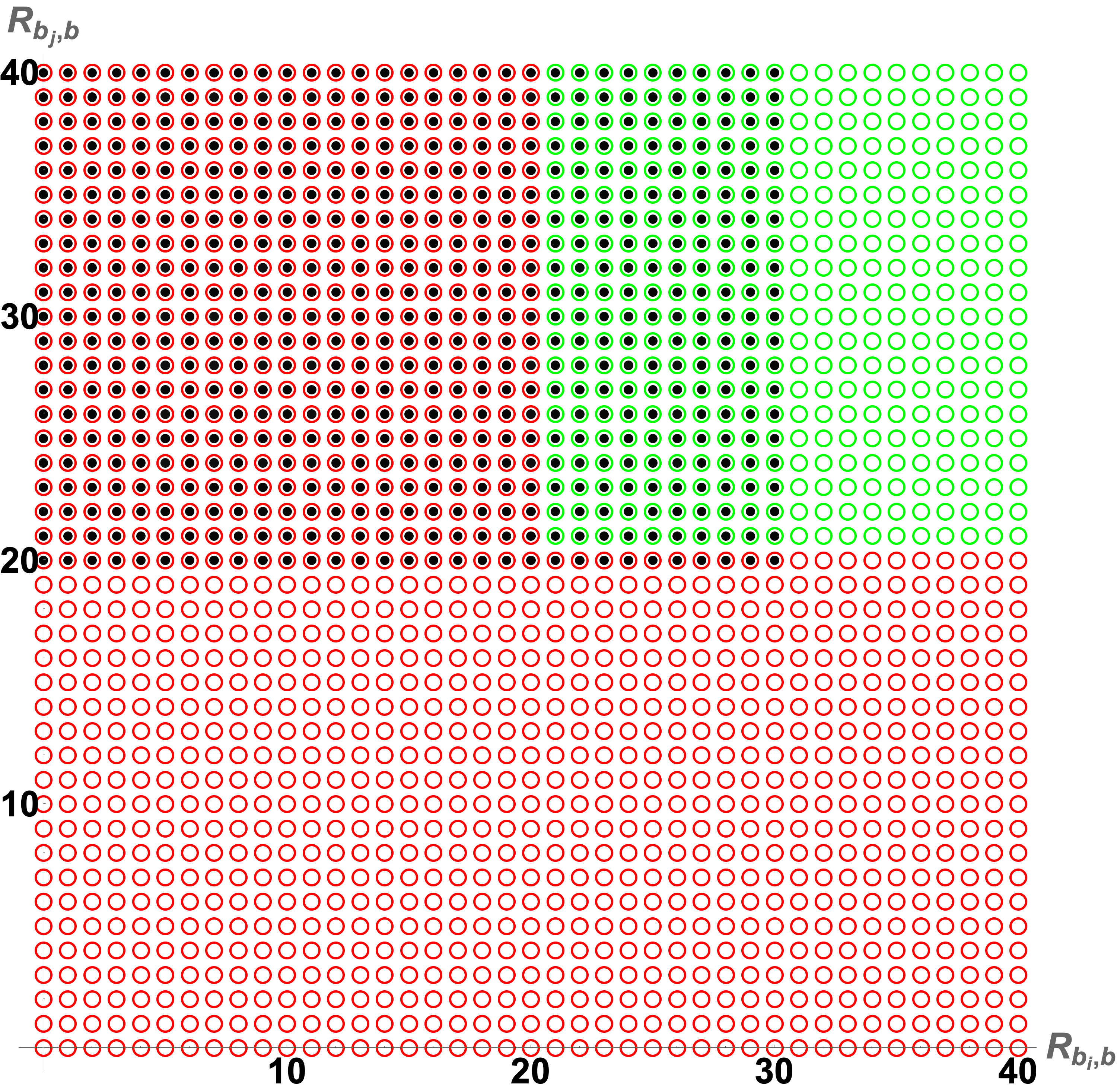}
         \label{fig:bLabelSquare}
     \end{subfigure}
     \hfill
        \caption{All the possible evaluations for a pair of binary classifiers
        for a $Q=100$ test assuming $\qa=60.$ The pair is working correctly for
        a label when its group evaluation lies on the green circles. Conversely,
        it is malfunctioning if it lies on the red circles. The computation of
        the possible evaluations was done for the hypothetical case of $\rai{1}=60$
        and $\rai{2}=20.$ For such a test result,
        there are no group evaluations for label \lbla that satisfy the safety
        specification of being better than 50\% on a label.}
        \label{fig:twoLabelSpacesForPair}
\end{figure}

The evaluation model for the trivial ensemble ($N=1$) is defined by the
observable response statistics \rai{i} and \rbi{i} - the number of times
a classifier $i$ gave a \lbla or \lblb response (our generic designation
of the two possible responses on each question in the test). For any finite
test of size $Q$, we can enumerate all the triples (\raia{i}, \rbib{i}, \qa)
that will contain the true evaluation of any responder These are the number of 
correct responses for each label, \raia{i}  and \rbib{i}. 

There are $1/6(Q+1)(Q+2)(Q+3)$ possible evaluations for a single classifier
in (\raia{i}, \rbib{i}, \qa) space (see Figure~\ref{fig:evaluation-cube}). 
No triplet outside this set can be considered a correct evaluation for any
classifier.
Enumerating the points in (\raia{i}, \rbib{i}, \qa) is not enough to fix an
evaluation model and how it is being used in its application context. 
There are two possible models, both with the same algebraic
logic. The first model is where we are directly evaluating binary classifiers.
In this model there is a fixed semantic equality between label responses to
questions. The second model is used when we are grading a binary response test
and there need not be any semantic equality between correct responses to questions.
Each has its own \emph{safety specification}.

In the case of evaluating binary classifiers we can formulate a \emph{safety specification}
such as,
\begin{equation}
    \pia{i} := \frac{\raia{i}}{\qa} > 50\%, \pib{i} := \frac{\rbib{i}}{Q - \qa} > 50\%,
    \label{eq:healthy-classifier-specification}
\end{equation}
a safe binary classifier is one that is better than 50\% on both labels. In the case
of binary response tests there is no relation between labels so the only meaningful
safety specification is that a responder was better than x per cent correct on the test,
\begin{equation}
    \pa \pia{i} + \pb \pib{i} > x\%,
\end{equation}
where \pa and \pb give the prevalence of \lbla and \lblb type questions on the test.
Neither of these prevalences has any semantic meaning outside the test.

In either of these two interpretations of a binary test, we can establish an
algebraic relation between observed responses and how correct the responder
was on the test,
\begin{align}
    \rai{i}  =& \raia{i} + (\raib{i} = \qb - \rbib{i}) \\
    \rbi{i}  =& (\rbia{i} = \qa - \raia{i}) + \rbib{i}.
    \label{eq:first-classifier-axiom}
\end{align}
These equations are complete but not independent. This follows from
the equation,
\begin{equation}
    Q = \rai{i} + \rbi{i}.
\end{equation}
Having observed \rai{i} or \rbi{i}, either of these equations defines
the same plane in (\raia{i}, \rbib{i}, \qa) space (Figure~\ref{fig:evaluation-cube}).
This reduces the number of possible evaluations for a classifier from $\mathcal{O}(Q^3)$
to $\mathcal{O}(Q^2).$ This variety and any equation that generates it defines
the $N=1$ axiom for our evaluation model of binary tests. All the members of
an ensemble of binary responders must obey it. This is shown in Figure~\ref{fig:evaluation-cube}
where we show the varieties associated with three LLMs that answered a binary response test.

\subsection{The N=2 construction}

Two binary classifiers have four possible decisions patterns (\raiaj{i}{j}, \raibj{i}{j},
\rbiaj{i}{j}, \rbibj{i}{j}) when we align
their decisions by item/question in the test. The counts for each of these patterns
can be expressed in terms of their individual (\raia{i}, \rbib{i}, \raia{j}, \rbib{j}) 
and joint correctness on the test
(\raiaja{i}{j}, \rbibjb{i}{j}). In the appendix we discuss how this generating set is
equivalent to  two copies of the single classifier
axiom (one for each classifier) and a new axiom for the pair,
\begin{equation}
    \raiaja{i}{j} + \rbibjb{i}{j} - \qa + \left( \rai{i} + \rai{j} \right) - \left( \raiaj{i}{j}\right) -
    \left( \raia{i} + \raia{j} \right)
\end{equation}
Similar to the case of the single classifier axiom, this has two versions. The one shown here
is the label \lbla version. Observing a pair classifiers introduces a new relation between test observables and
statistics of correctness on it. And as we would expect, each member of the pair still
obeys the single classifier axiom. This confirms what we would suspect intuitively since
the single classifier axiom only involves quantities related to one classifier alone.
It may seem overkill to claim these relations are ``axioms'' for the logic of unsupervised
evaluations of binary classifiers. This topic is discussed further in the appendix in the context
of detecting corrupted or spoofed test summaries thereby showing the connection of
this logic with trustworthiness.

\section{Verifying at least one classifier is malfunctioning}

The axioms in the previous section are, themselves, verifiers of group evaluations.
Given observed responses, we can ask - what group evaluations satisfy them? Any
evaluation algorithm that returned evaluations that violated them would be
certifiably wrong. This ability to prove that a group evaluation is logically
consistent with test response statistics can be used to create a logical alarm
for misaligned classifiers. We will do so using the semantic interpretation of
the binary test. Our arbitrary \emph{safety specification} will be that all
classifiers are required to be better than 50\% on each label 
(Equations~\ref{eq:healthy-classifier-specification}). In general, any range is
possible as will become clear from the geometrical nature of the algorithm.

The general idea of the alarm is that all the classifiers in an ensemble must
satisfy the single classifier axiom. We do not know the actual value of \qa 
in a fully unsupervised setting. But we know that the true value must be an
integer between 0 and $Q.$ At each fixed \qa value, the first classifier axiom
defines a line establishing a dependency relation between \raia{i} and \rbib{i}
- the plane defined by the axiom is intersected by the horizontal plane at the assumed
\qa value. This allows us to define the only pair evaluations consistent with test
responses at that value of \qa. If no group evaluation satisfies the safety
specification, the system is malfunctioning at the assumed \qa value. We can
continue this procedure for all possible values of \qa and if at each assumed
setting the ensemble fails the safety specification, we know that at least one
classifier is malfunctioning.

\subsection{The rectangle of logically consistent label evaluations at fixed \qa}

At fixed \qa we can compare pairs of classifiers ($i, j$) by finding the set of possible
evaluations in two separate spaces, one for each label. For the \lbla space,
we use the variables (\raia{i}, \raia{j}). The \lblb space is defined by the
variables (\rbib{i}, \rbib{j}). Since \qa is fixed, the set of possible evaluations
in each space defines a square of points (Figure 2). For label \lbla, the square
is an integer lattice going from 0 to \qa. And for label \lblb, the square is
an integer lattice from 0 to $Q - \qa.$ These are the possible group evaluations
for the pair before we use the axioms. 

The single classifier axiom restricts the set of possible correct responses in
each of the label spaces.
We illustrate how with the label \lbla. The single classifier axiom can
be written expressing \rbib{i} in terms of $Q$, \qa, \rai{i}, and \raia{i} as
\begin{equation}
    \rbib{i} = Q - \qa - \rai{i} + \raia{i}.
\end{equation}
But we know that the number of correct \lblb responses, \rbib{i}, must be between 0 and
\qb for any given classifier $i$
\begin{equation}
    0 \leq Q - \qa - \rai{i} + \raia{i} \leq Q - \qa.
\end{equation}
This equation thus defines
the values of \raia{i} that are consistent with the single classifier
axiom at the assumed \qa value. 
Since we can do this for both members of a pair, the subset of group
evaluations is a rectangle in \lbla space. A similar argument can
be used to define the logically consistent rectangle in \lblb space
starting with the \lblb version of the single classifier axiom,
\begin{equation}
    0 \leq \qa - \rbi{i} + \rbib{i} \leq \qa.
\end{equation}
In general, we would be able to use the single classifier axiom for
an ensemble of any size, in each case there would be two label
spaces, each of dimension $N$, where can find a cuboid defined
by the application of the axiom for each classifier in both
label spaces.

\subsection{Testing at all possible values of \qa}

In a fully unsupervised setting the value of \qa, itself, is not known.
But its value is finite and we know that it lies between 0 and $Q.$
It thus becomes possible to test if the ensemble violates the safety
specification at all assumed values of \qa - if not we know that at least
one member of the ensemble is malfunctioning. An example is shown in Figure 2
and discussed in detail in the Appendix.

This logical argument cannot tells us under what conditions -
other than possible values of the evaluation sketch - this detection becomes
possible. In other words, what constitutes enough of a difference in agreement
to trigger the alarm remains an engineering problem. This is similar to how
the sensitivity of fire or gas alarms are determined by their application context.
What is notable here is that this alarm is purely based on logical consistency
of test responses. A specific example is detailed in the appendix and an
illustrative example is shown in Figure~\ref{fig:twoLabelSpacesForPair}.

The alarm is not foolproof. If all members of an ensemble are malfunctioning
in the same manner, this algorithm cannot detect anything wrong. The algorithm can
only detect that the classifiers are misaligned. In this regard, engineering use
of the alarm should follow a \emph{defense in depth} design - creating ensembles
large enough that the failure of a few members is more likely than all of them at once.
Additionally, in semi-supervised evaluation settings, we may have side information that can ground
the alarm. For example, the range of possible \qa values may be known. For misaligned
classifiers, correctly satisfying the safety specification occurs at \qa values that
are not the true one.

\section{Discussion and previous work}

We have constructed a series of evaluation models for binary evaluations
of $N$ noisy agents. These allow us to identify a nested set of axioms
(all singletons, all pairs, etc...) that must be satisfied by any
binary evaluation. These axioms define the set of all group evaluations
consistent with just their observed responses. The axioms serve as
verifiers and can reject incorrect evaluations. They can also be used
to detect ensembles that violate safety specifications expressible in
terms of statistics of correctness on binary tests.
This logic for unsupervised evaluation is therefore a concrete example
of how formal verification methods can be used to help us monitor
noisy agents. As we can see, it is much easier to
formulate and formalize evaluation models than world models.
 Another
advantage of a logic of fully unsupervised evaluation is that we can apply
it to circumstances where an \emph{answer key} exists but we suspect it
is wrong or has been spoofed - it is a tool for checking trustworthiness.

Most work in the ML/AI literature on unsupervised evaluation is about
creating evaluation algorithms. The seminal paper
by Dawid and Skeene \cite{Dawid79} used a likelihood minimized
with the EM algorith to estimate the accuracy of doctors reviewing medical charts
for diagnosis. Subsequently, it has received many probabilistic
treatments at this conference and others. One stream could be characterized
as the Bayesian approach \cite{Raykar2010, Wauthier2011, Zhou2012, Liu2012, Zhang2014}.
A spectral approach was initiated by Parisi et al \cite{Parisi1253} and
further developed by Jaffe et al \cite{Jaffe2015, Jaffe2016}. Evaluation is an
abstract task that can occur anywhere in the ML cycle. Unsupervised evaluation
issues occur during supervised training of classifiers \cite{Su2024}.
The work closest to the algebraic, logical approach taken here is by
Platanios et al and their agreement equations \cite{Platanios2014, Platanios2016}. 
They correctly noted the purely algebraic and logical basis for their work and that of others. 
Agreements, however, are just 2 out of $2^N$ events so the agreement equations do
not form a complete generating set and therefore cannot be used for verification.
The Platanios solution to evaluation for error-independent classifiers \cite{Platanios2014}
is incorrect as discussed in the appendix.

The use of algebraic geometry in statistics was pioneered by Pistone et al
\cite{Pistone}. They were not concerned with sample statistics as we are
here, but rather on experimental design and inference problems with distributions.
Using sample statistics for evaluation has many similarities to data streaming
algorithms and error-correcting codes as discussed further in the appendix.

\section{Limitations and societal impacts}
Formalization of verifying measurements cannot resolve the problem
of interpreting them. For that one requires a \emph{world model}.
Formalization of unsupervised evaluation is possible because it deals
with \emph{evaluation models} of sample statistics of an evaluation.
As such, it cannot tells us anything about future or past values for
those statistics. That is the job of \emph{evaluation models} that
incorporate probability assumptions or other domain knowledge rules.
An analogy with safety engineering in other realms may help the reader
circumscribe properly the power and limitations of any logic of unsupervised
evaluation.

Thermometers and smoke detectors are used as alarm components within
safety frameworks. A thermometer can be used to alarm the on-board computer
that a car engine is overheating. A smoke detector can bring attention to
a possible fire. Neither the thermometer nor the smoke alarm have much
\emph{intelligence} of their own. They cannot tell you what causes the
over heating or smoke. Nor can they diagnose how to fix the problem.
Logics of unsupervised evaluation can serve a similar role within
safety frameworks of noisy agents. Doctors use thermometers to
help keep patients safe.

Responsible use of any measurement methodology requires that we understand
the effects of over reliance on it. This has been noted in the AI safety
literature. For example, Dalrymple et al's \cite{Dalrymple2024} mention of Goode's Law.
Or even the misalignment of single measures with human values \cite{Wang2022}.
Formalization also can lull its users into believing all its well. We
see here how misguided that can be in how the logical alarm is constructed.
It can never prove that all the classifiers are working correctly.
It can only detect when they are not. That it can certify with logical
certainty. But the logical converse is not possible. All true group evaluations
where the ensemble members are behaving roughly similar, whether correctly
or not, will not trigger the logical alarm presented here.

One positive societal benefit of this formalization follows directly from the
previous statement. It is a direct demonstration of the utility of noisy
agents when performing any difficult task. It is only when agents disagree
that we can use their own decisions to self-evaluate them. Even in
cases where there is a high performing agent, whether human or
robotic, an ensemble of noisy, weaker agents can be used to supervise
it via this evaluation logic.

A second benefit from logics of unsupervised evaluation is the role they
play in the economic problems related to principal/agent interactions.
As is discussed in the Appendix, exact, fully algebraic evaluation is
possible when the noisy agents are error independent on a test.
Even in that exact solution, two possible group evaluations exist.
There always has to be some principal that establishes the correct
one. Supervision is always necessary even for something so simple
as binary response evaluations. But this logic makes it much
easier since it serves as sieve for possible evaluations. Any 
Bayesian calculation that computes the number of questions 
that need to be ground checked to establish a desired range
of possible evaluations would be able to start with a set
at least $1/Q$ smaller than the fully ignorant set (all
possible evaluations). This could be as small as 1 for certain
test results. If we detect that one of the classifiers
is 100\% or 0\% correct on both labels, we would just need to check
one question to ascertain which evaluation was correct. This would
simultaneously ground the evaluation of all the other members of the
ensemble in the case of error-independent test results.

Finally, some researchers are concerned that the problem of super-alignment -
being able to supervise agents smarter than us - is fundamentally unsolvable.
If that was the case, it would be the first technology for which we cannot
build controls than are simpler and less intelligent than the systems they
control. Special case solutions like the error-independent evaluation model
allow us to engineer systems that we can evaluate on any binary test -
the algorithms are devoid of any semantics of the world. In this way, it
is a tool like the steam governor controls runaway locomotives, the
car thermometer alarms us about overheating engines, and the smoke alarm warns
us about possible fires. If fires are not usually controlled by building
bigger fires, so we should consider that less-intelligent mechanisms are
an integral component of any safe and trustworthy system.

\begin{ack}
Andr\'es Corrada-Emmanuel was the lead author and drafted the manuscript.
He gratefully acknowledges the financial support of the U.S.\ Naval
Research Laboratory as Summer Research Fellow.
Ilya Parker and Ramesh Bharadwaj are contributing authors and are
responsible for recognizing that the algebraic formalism used to obtain
the error independent solution, the subject of a 2010 U.S.\ patent,
could be interpreted as a logic and make formal verification of
unsupervised evaluations possible.
\end{ack}

\bibliography{main}{}
\bibliographystyle{plain}

\appendix

\section{Appendix / supplemental material}

The machinery for formalizing the logic of unsupervised evaluation
already exists in the field of Algebraic Geometry (AG). Theorem provers
in geometry are another example of how AG is used in formal verification
(Chapter 10 in Cox \cite{Cox}). This appendix is written in a gentler
style that is going to sketch the AG proofs. This may help the reader
become familiar with AG if this is their first encounter.

\subsection{Definitions and the set of all possible evaluations for tests of size $Q$}

The paper discussed evaluation models in \emph{response space} or 
\rspace. Researchers
in ML and AI usually discuss evaluation in \emph{percentage space} or
\pspace. This
section provides definitions for all the relevant statistics discussed in the
paper in both spaces.

\begin{table*}\centering
\ars{1.3}
\begin{tabular}{@{}lcc@{}}\toprule
 \phantom{abc} & test statistics & classifier statistics \\
R space: integers\\
\cmidrule{1-1}
observables & Q & \rai{i}, \rbi{i} \\
unobservables & \qa, \qb & \raia{i}, \rbib{i} \\
\addlinespace[0.8em]
P space: rationals\\
\cmidrule{1-1}
observables & Q & $\fai{i} = \frac{\rai{i}}{Q}$, $\fbi{i} = \frac{\rbi{i}}{Q}$ \\
\addlinespace[0.5em]
unobservables & $\pa = \frac{\qa}{Q}, \pb = \frac{\qb}{Q}$ & $\pia{i} = \frac{\raia{i}}{\qa}, \pib{i} = \frac{\rbib{i}}{\qb}$ \\
\addlinespace[0.5em]
\bottomrule
\end{tabular}
\caption{Statistics of an evaluation given to a single binary classifier}
\label{tbl:statistics}
\end{table*}

\subsubsection{\rspace definitions}
The \emph{observable statistics}
are,
\begin{itemize}
    \item $Q$: number of items/questions in the evaluation.
    \item \rai{i}: number of times classifier $i$ responded \lbla.
    \item \rbi{i}: number of times classifier $i$ responded \lblb.
\end{itemize}
The \emph{unobservable statistics} are,
\begin{itemize}
    \item \qa: number of items/questions with correct response \lbla.
    \item \qb: number of items/questions with correct response \lblb.
    \item \raia{i}: number of times classifier $i$ responded correctly to \lbla questions.
    \item \rbib{i}: number of times classifier $i$ responded correctly to \lblb questions.
\end{itemize}

\subsubsection{\pspace definitions}

The \emph{observable statistics}
are,
\begin{itemize}
    \item $Q$: number of items/questions in the evaluation.
    \item $\fai{i}=\rai{i}/Q$: percentage of times classifier $i$ responded \lbla.
    \item $\fbi{i}=\rbi{i}/Q$: percentage of times classifier $i$ responded \lblb.
\end{itemize}
The \emph{unobservable statistics} are,
\begin{itemize}
    \item $\pa=\qa/Q$: percentage of correct \lbla responses.
    \item $\pb=\qb/Q$: percentage of correct \lblb responses.
    \item $\pia{i}=\raia{i}/\qa$: percentage of correct \lbla responses.
    \item $\pib{i}=\rbib{i}/\qb$: percentage of correct \lblb responses.
\end{itemize}

\subsubsection{The set of all possible single binary classifier evaluations of size $Q$}

Evaluation models are much easier than world models. We can enumerate all the possible
evaluations for a single binary classifier in \rspace, (\raia{i}, \rbib{i}, \qa).
The following algorithm generates all of them for a test of size $Q$,

\begin{algorithm}[H]
\SetAlgoLined
\KwResult{All possible evaluations, $(\raia{i}, \rbib{i}, \qa)$, 
for a test with $Q$ questions.}
evaluations $\leftarrow [\,]$ \tcp{Initialize possible evaluations with the empty list.}
\For{$\qa \leftarrow 0$ \KwTo $Q$}{
  \For{$\raia{i} \leftarrow 0$ \KwTo \qa}{
    \For{$\rbib{i} \leftarrow 0$ \KwTo $Q - \qa$}{
      evaluations.append($(\raia{i}, \rbib{i}, \qa)$)
    }
  }
}
\end{algorithm}

\subsubsection{The evaluation ideals in \rspace and \pspace}

\begin{figure}
     \centering
     \begin{subfigure}[b]{0.45\textwidth}
         \centering
         \includegraphics[width=\textwidth]{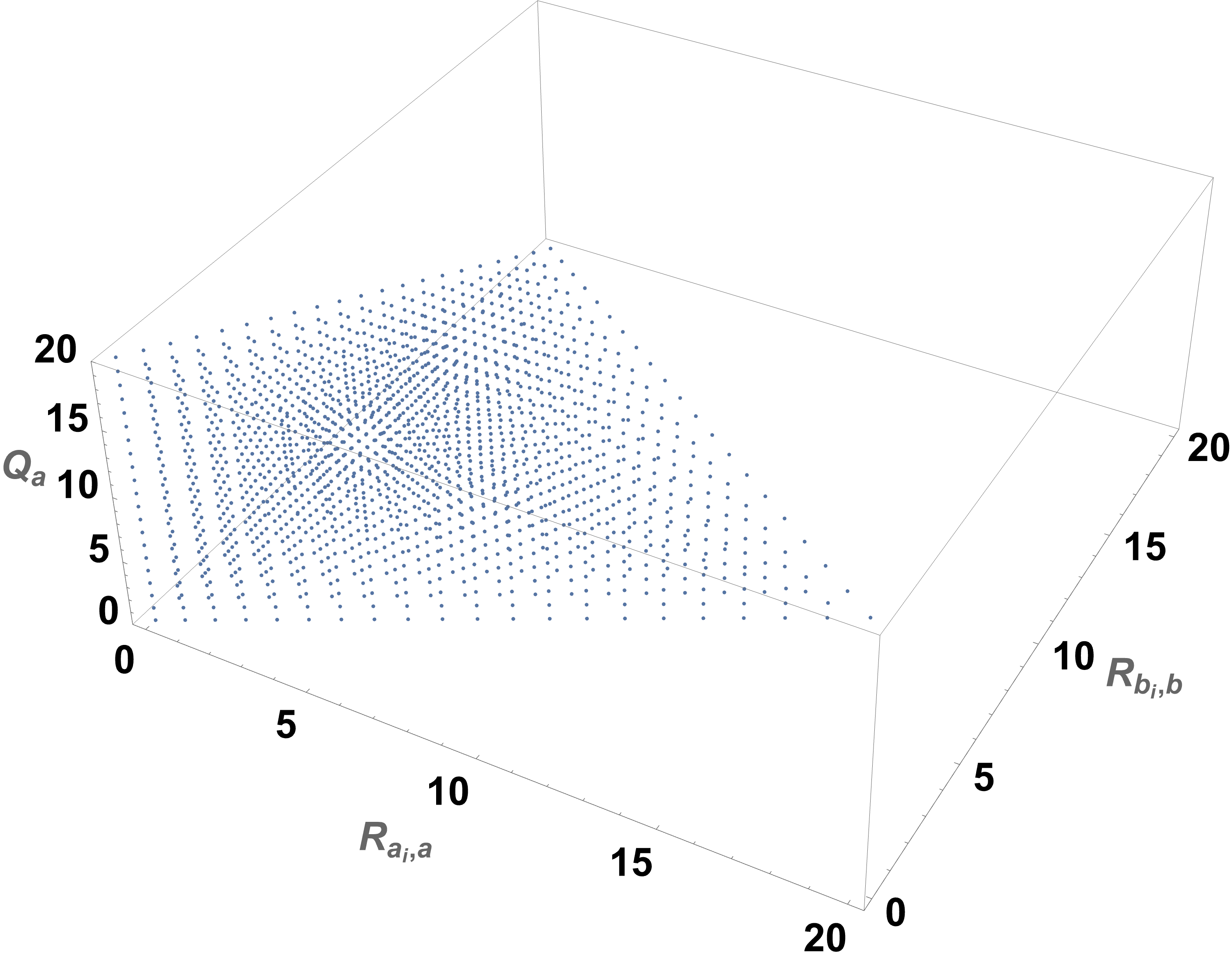}
         \label{fig:rSpaceAllPossible}
     \end{subfigure}%
     \hspace{2pt}
     \begin{subfigure}[b]{0.45\textwidth}
         \centering
         \includegraphics[width=\textwidth]{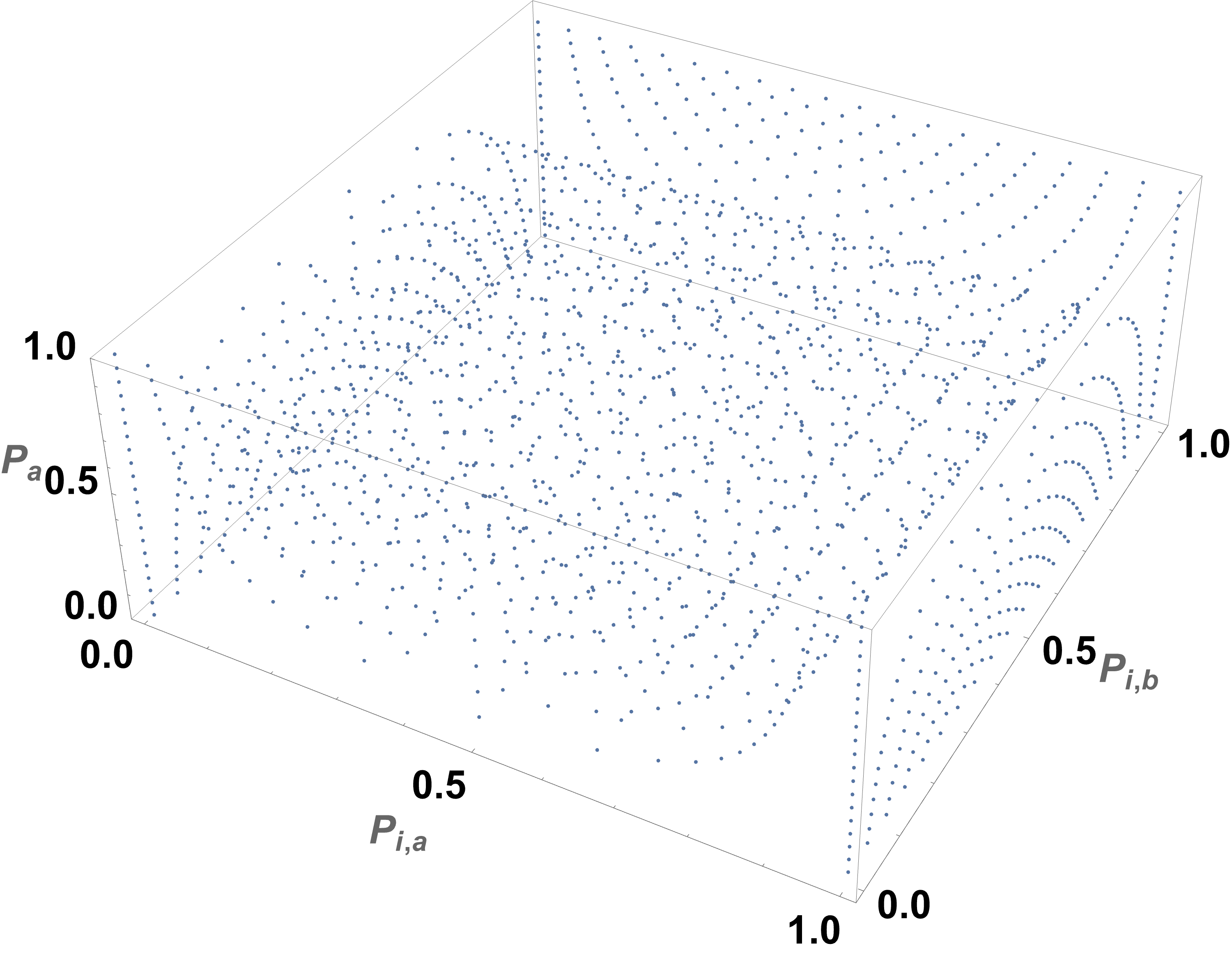}
         \label{fig:pSpaceAllPossible}
     \end{subfigure}
     \hfill
        \caption{All the possible single classifier test summaries for
        a $Q=20$ test. The left figure shows them in \rspace, the right
        one in \pspace. Note the difference in the geometry of each set.}
        \label{fig:twoSpaces}
\end{figure}

The set of all possible points for the single classifier test summary looks different
in each space. Figure~\ref{fig:twoSpaces} shows an example for a $Q = 20$
test. After we have observed a classifier responses, we can invoke the single
classifier axiom which can be written in each space,
\begin{align}
    \pa (\pia{i} - \fai{i}) & - \pb (\pib{i} - \fbi{i}) \\
    (Q \raia{i} - \qa \rai{i}) & - (Q \rbib{i} - \qb \rbi{i}).
\end{align}
The \rspace version of the axiom can be manipulated into the forms necessary for
the construction of the consistent cuboid by using the identity,
\begin{equation}
    Q = \rai{i} + \rbi{i}.
\end{equation}

\subsection{Construction of the N=2 axiom}

There are two approaches to deriving the $N=2$ axiom, one in
\rspace the other in \pspace. We first show the hardest approach, using \pspace, because
it highlights test statistics related to decision error correlations between
the members of an ensemble. The second approach, using \rspace, is direct
and readily generalized to any ensemble of size $N.$

\subsubsection{The $N=2$ generating set in \pspace and its axioms}

In \pspace one uses sample \emph{frequencies} of all the possible
$2^N$ patterns, the observed count of the patterns divided by
the size of the test. This is what guarantees their completeness - there
are no other patterns that we do not know about.
For ensembles of size $N=2$ with classifiers, $\left\{i,j\right\},$
that completeness is given by,
\begin{equation}
    \faaij{i}{j} + \fabij{i}{j} + \fbaij{i}{j} + \fbbij{i}{j} = 1.
\end{equation}
For each of the 4 decision patterns, we can write a polynomial
of variables in \pspace,
\begin{align*}
    \faaij{i}{j} &= \pa \left( \prsaa{i} \prsaa{j} + \corratwo{i}{j} \right) + 
    \pb \left( \left( 1 - \prsbb{i} \right) \left( 1 - \prsbb{j} \right) + \corrbtwo{i}{j} \right)\\
    \fabij{i}{j} &= \pa \left( \prsaa{i} \left(1 - \prsaa{j} \right) - \corratwo{i}{j} \right) + 
      \pb \left( \left( 1 - \prsbb{i} \right) \prsbb{j} - \corrbtwo{i}{j} \right)\\
    \fbaij{i}{j} &= \pa \left( \left(1 - \prsaa{i} \right) \prsaa{j}  - \corratwo{i}{j} \right) + 
      \pb \left( \prsbb{i} \left( 1 - \prsbb{j} \right) - \corrbtwo{i}{j}\right)\\
    \fbbij{i}{j} &= \pa \left( \left(1 - \prsaa{i} \right) \left(1 - \prsaa{j} \right)  + \corratwo{i}{j} \right) + 
      \pb \left(  \prsbb{i} \prsbb{j} + \corrbtwo{i}{j}\right)
\end{align*}
Since all the observable decision frequencies are on the left and all the statistics
of correctness are on the right, this is a map from \pspace to what we can call
\fspace. This justifies calling them a \emph{generating set} - given the statistics
of correctness for one's chosen evaluation model, they generate the decision frequencies
we have seen on the test.

Although more complicated, the \pspace generating set for the pair ensemble introduces
new sample statistics of correctness that are needed to generate \emph{all} the possible
decision patterns we could see from a binary evaluation. Sample statistics are needed to
represent the decision correlations between the members of the ensemble. We only
need two, one for each label, for binary evaluations -  \corratwo{i}{j} and \corrbtwo{i}{j}. 
In general we need $R^R - R$ for
classifications/tests with $R$ labels/responses.
They are defined as,
\begin{align}
    \label{eq:pair-complete-set}
    \corratwo{i}{j} &:= \frac{\raiaj{i}{j}}{\qa} - \frac{\raia{i}}{\qa} \frac{\raia{j}}{\qa} \\
    \corrbtwo{i}{j} &:= \frac{\rbibj{i}{j}}{\qb} - \frac{\rbib{i}}{\qb} \frac{\rbib{j}}{\qb}
\end{align}
These are algebraic definitions. They define a different notion of \emph{independence}
from that of \emph{distributional independence} that is used in probability theory.
Notions of \emph{independence} occur in many different contexts. We have just
remarked on two different ones. In vector theory one talks of 
\emph{independent vectors}. Unfortunately, there is no notion of \emph{independence}
for polynomials similar to \emph{vector independence}. But there are
ways of rewriting the generating sets that disentangle these new error correlation
variables for any ensemble of size $N.$

In algebraic geometry, the two central objects of study are algebraic ideals
(infinite sets of polynomials) and algebraic varieties (sets of points in
the space of the polynomial variables that zero out all members of the ideal).
The algebraic varieties can be finite as they are in binary evaluations.
An example is shown in Figure~\ref{fig:twoSpaces}. But algebraic ideals
are infinite. Hilbert famously proved that all algebraic ideals are
finitely generated - there always exist a finite set of polynomials
that generate all its members.

It is tempting to think that because we always have a finite set
of polynomials that generate all the observations in a space
of dimension equal to its size that we could think of them
as ``vectors'' in polynomial space. We cannot. For one, there
exist other generating sets for any given ideal that could
be larger or smaller. There is no fixed notion of ``dimension''.
Nonetheless, there are ways of writing generating sets that are
better than others.

\emph{Gr\"oebner bases} were first described in the 1960s and
are central to the field of computational algebraic geometry.
Finding a Gr\"oebner basis is done using Buchberger's algorithm
\cite{Cox}. There is not one Gr\"oebner basis but many for
a given finite generating set as we have for the evaluation
model above. One of those choices has 6 polynomials for
the generating set. But some of them, although different
in algebraic form, can be transformed into each other.
Those are the \emph{axioms} of an evaluation model. For
the $N=2$ ensemble we recover the single classifier axiom,
one for each of the classifiers.
\begin{gather}
    \pa \left(\pia{i} - \fai{i}\right) - \pb \left(  \pib{i} - \fbi{i} \right) \\
    \pa \left(\pia{j} - \fai{j}\right) - \pb \left(  \pib{j} - \fbi{j} \right).
\end{gather}
In addition, we recover a pair axiom that can be written as different expressions but we
can transform into each other by using the single classifier axioms. Both forms are
informative about the limitations of inferring evaluations from decision patterns,
\begin{multline}
\left(\pia{i} + \pib{i} - 1\right) \left(\pia{i} - \fai{i}\right) \left(P_{j, b} - f_{b_j}\right)  \\
    + \left( \Gamma_{i, j, b} - \Delta_{i,j} \right) \left(P_{i, a} - f_{a_i}\right)  \\
    + \left( \Gamma_{i, j, a} - \Delta_{i,j} \right) \left(P_{i, b} - f_{b_i}\right),
\end{multline}
and
\begin{multline}
\left(P_{j, a} + P_{j, b} - 1\right) \left(P_{i, a} - f_{a_i}\right) \left(P_{j, b} - f_{b_j}\right)  \\
  + \left( \Gamma_{i, j, b} - \Delta_{i,j} \right) \left(P_{j, a} - f_{a_j}\right)  \\
  + \left( \Gamma_{i, j, a} - \Delta_{i,j} \right) \left(P_{j, b} - f_{b_j}\right) 
\end{multline}
There are ``blindspots'' that zero out components of these polynomials and thereby make them less useful in
restricting the set of logically consistent evaluations. The terms like,
\begin{equation}
\left(P_{j, a} + P_{j, b} - 1\right),
\end{equation}
become zero when, for example, the classifier just guesses a fix proportion of the labels.
Irrespective of how it happened, if the evaluation statistics lie on this line, the
set of logically consistent evaluations will increase. Likewise, there is a point
on this line defined by,
\begin{align}
    \pia{i} & = \fai{i} \\
    \pib{i} & = \fbi{i}
\end{align}
that makes the logically consistent set equal to the set of all possible evaluations
before the test results are observed.

The term $\Delta_{i,j}$ is derived from the agreement/disagreement frequencies. It
is defined as either,
\begin{align}
    \Delta_{i,j} & := \faaij{i}{j} - \fai{i} \fai{j} \\
                 & := \fbbij{i}{j} - \fbi{i} \fbi{j},
\end{align}
since both can be shown to be equivalent using the completeness of pair
decision patterns.
The value of $\Delta_{i,j}$, also identifies another way the axiom can
become a simpler polynomial. We will have more to say about this in the
section discussing the error independent solution.

\subsubsection{The $N=2$ axiom in \rspace}

The generating set in \rspace can be derived from that in \pspace. 
It is shown here so the reader can compare it with the \pspace 
representation.It
generates the (\raiaj{i}{j}, \raibj{i}{j}, \rbiaj{i}{j}, \rbibj{i}{j})
test summaries,
\begin{align}
    \raiaj{i}{j} & = \raiaja{i}{j} + \left( \qb - \rbib{i} - \rbib{j} + \rbibjb{i}{j} \right) \\
    \raibj{i}{j} & = \left( \raia{i} - \raiaja{i}{j} \right) + \left( \rbib{j} - \rbibjb{i}{j} \right) \\
    \rbiaj{i}{j} & = \left( \raia{j} - \raiaja{i}{j} \right) + \left( \rbib{i} - \rbibjb{i}{j} \right) \\
    \rbibj{i}{j} & = \left( \qa - \raia{i} \raia{j} + \raiaja{i}{j} \right) + \rbibjb{i}{j}
\end{align}
The reader can see this \rspace representation is not as symmetric as the
\pspace representation. It also does not make clear what is the role of
error correlation between the members of the pair.
Instead of using this generating set to find the $N = 2$ axiom in \rspace,
we are just going to construct it directly. There are two equivalent
constructions as we have mentioned before. We will detail the label \lbla
construction. The construction for label \lblb merely changes the label.

We want to derive an expression for the quantity \rbibjb{i}{j}, the
number of observed counts were both responders answered \lblb to
\lblb type questions. To do that, we write it as \qb minus the
number of times each alone gave an \lbla response incorrectly
minus the number of times they both said \lbla incorrectly.
The number of times related to both making the mistake alone
gives us our first terms,
\begin{equation}
    \rbibjb{i}{j} = \qb - (\rai{i} + \rai{j}) + (\raia{i} + \raia{j}) \ldots
\end{equation}
But this under counts by one each time they both got it wrong so
we correct with the joint decision observed and correct counts,
\begin{equation}
     \rbibjb{i}{j} = \qb - \left( (\rai{i} + \rai{j}) - (\raia{i} + \raia{j})\right) 
     + \left( \raiaj{i}{j} - \raiaja{i}{j} \right)
\end{equation}

\subsubsection{Using the $N=2$ axiom to further restrict logically consistent evaluations}
The pair evaluation axiom allows us to restrict further the set of member group
evaluations consistent with their aligned responses on a test. We can rearrange
the axiom to give us an expression for the sum of their jointly correct
response counts,
\begin{equation}
    \raiaja{i}{j} + \rbibjb{i}{j}
\end{equation}
This expression has two equivalent formulations, just like in the single classifier
axiom case. These are,
\begin{gather}
    \qb - (\rai{i} + \rai{j}) + \raiaj{i}{j} + (\raia{i} + \raia{j})\\
    \qa - (\rbi{i} + \rbi{j}) + \rbibj{i}{j} + (\rbib{i} + \rbib{j})
\end{gather}
These expressions can be used to restrict further the possible group
evaluations for a pair. For example, in label \lbla space, small
values of the sum $\raia{i} + \raia{j}$ may not be enough to cause
this expression to be zero or positive. Since the sum of correct
joint responses can never be below zero, this would prove that the
individual correct counts must be larger at the assumed value of
\qa. An example of how this restriction "cuts the corners" of
the admissible pair evaluation rectangle is given in the section
discussing the BIG-Bench-Mistake multistep arithmetic evaluation.

\subsection{LLMs grading other LLMs: an evaluation using the BIG-Bench-Mistake
multistep-arithmetic CoT task}

Completeness in a logic of unsupervised evaluation has an important practical
use - it terminates evaluation chains. This is being demonstrated in this paper
by building a logical alarm that can certify that at least one ensemble member
is failing the safety specification. In this section we illustrate this 
use for the formalism by discussing a binary evaluation used when three LLMs 
(Claude, Mistral, and GPT4) graded a PaLM2 LLM that
had been given the multistep-arithmetic task from the BIG-Bench-Mistake
dataset \cite{Tyen2024}.

Tyen at al \cite{Tyen2024} created the dataset to study the reasoning
abilities of LLMs. The multistep-arithmetic task consists of 300 problems
of the form,
\begin{equation}
    (((-9 - 5 - 0) - (4 + 3 + -5)) - ((3 * 4 * 5) * (7 - -7 * 4))) = ?
\end{equation}
A full evaluation of the reasoning ability of LLMs is much more than
a binary evaluation. But such an evaluation is possible and could detect
that the LLMs supervising an LLM are, themselves, not doing their
work correctly. That was done by asking a grading LLM to make a binary
decision on each answer provided by the PaLM2 LLM - "Is this answer
correct?"

\begin{table*}\centering
\ars{1.3}
\begin{tabular}{@{}lcc@{}}\toprule
 pattern & \lbla & \lblb \\
\cmidrule{2-2} \cmidrule{3-3}
(\lbla, \lbla, \lbla) & 12 & 0 \\
(\lbla, \lbla, \lblb) & 0 & 0 \\
(\lbla, \lblb, \lbla) & 113 & 8 \\
(\lblb, \lbla, \lbla) & 14 & 0 \\
\addlinespace[0.5em]
(\lblb, \lblb, \lbla) & 72 & 15 \\
(\lblb, \lbla, \lblb) & 0 & 1 \\
(\lbla, \lblb, \lblb) & 11 & 2 \\
(\lblb, \lblb, \lblb) & 15 & 18 \\
\addlinespace[0.5em]
\bottomrule
\end{tabular}
\caption{Grading agreements and disagreements between three LLMs (Claude Haiku, Mistral
Large, GPT4-Turbo) that checked the answers of a PaLM2 LLM doing the multistep
arithmetic task in the BIG-Bench-Mistake dataset. The
label \lbla means "incorrect", label \lblb means "correct."}
\label{tbl:grading-llms}
\end{table*}

\begin{figure}
  \centering
  \includegraphics[width=0.75\textwidth]{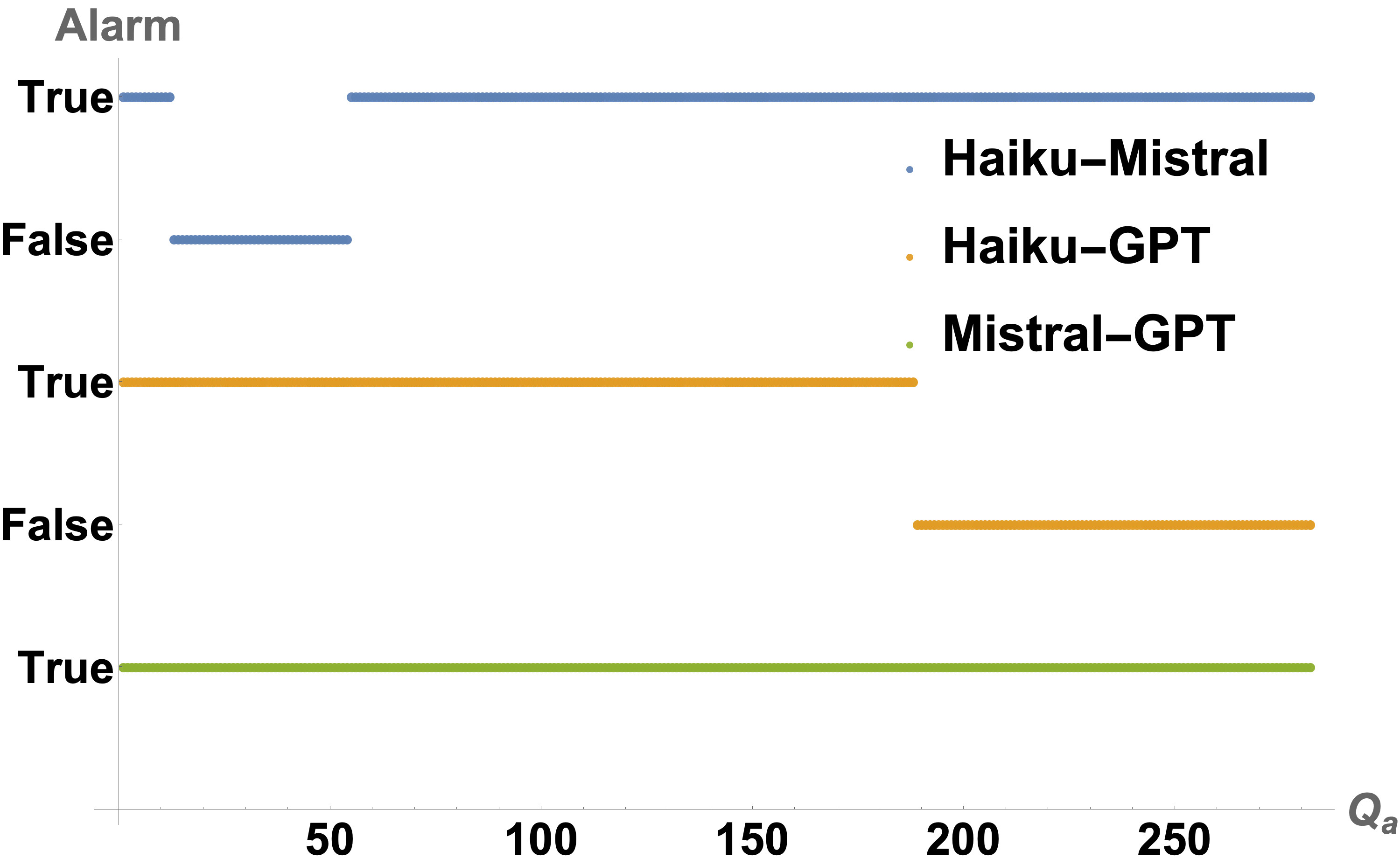}
  \caption{Logical traces for the misalignment alarm based on the
  single classifier axiom for binary classifiers. Three LLMs 
  graded a fourth one completing the multistep-arithmetic task 
  in the BIG-Bench-Mistake dataset. One pair, the Mistral-Large
  and GPT4-Turbo LLMs, disagreed enough to violate the safety
  specification at all assumed values \qa.}
  \label{fig:logical-traces}
\end{figure}

The evaluation detailed here occurred when we used three commercially
available LLMs - Claude Haiku by Anthropic, Mistral-Large by Mistral AI, and 
GPT4 Turbo by Open AI. We summarize the grading agreements and disagreements by the
LLMs by true label in Table~\ref{tbl:grading-llms}. By marginalizing the counts
for each grader, we obtain three pairs of inequalities that form the core of
the logical alarm. For the particular evaluation in Table~\ref{tbl:grading-llms},
these inequalities are,
\begin{align}
    0 & \leq (Q - \qa) - 146 + \raia{1} \leq (Q - \qa) \\
    0 & \leq \qa - 135 + \rbib{1} \leq \qa \\
    0 & \leq (Q - \qa) - 27 + \raia{2} \leq (Q - \qa) \\
    0 & \leq \qa - 254 + \rbib{2} \leq \qa \\
    0 & \leq (Q - \qa) - 234 + \raia{3} \leq (Q - \qa) \\
    0 & \leq \qa - 47 + \rbib{3} \leq \qa \\
\end{align}
These inequalities allow us to define cuboids for ensembles of $N > 1$ as explained
in the paper. At each fixed value of \qa we can construct the set of evaluations
logically consistent with this axiom and test if it fails the safety specification
for any of the classifiers. This is done for the three possible pairs and are
shown in Figure~\ref{fig:logical-traces} as the traces of the logical test at each fixed 
\qa value.

The only pair triggering the alarm for all values of \qa is the Mistral-Large and GPT4-Turbo
pair. Both are failing the safety specification by being less than 50\% on one of the labels.
This example highlights that the alarm is not triggered by correctness of any of its
members since the logical algorithm does not use the answer key or impute one other
than by some statistic like \qa.

\begin{figure}
  \centering
  \includegraphics[width=0.75\textwidth]{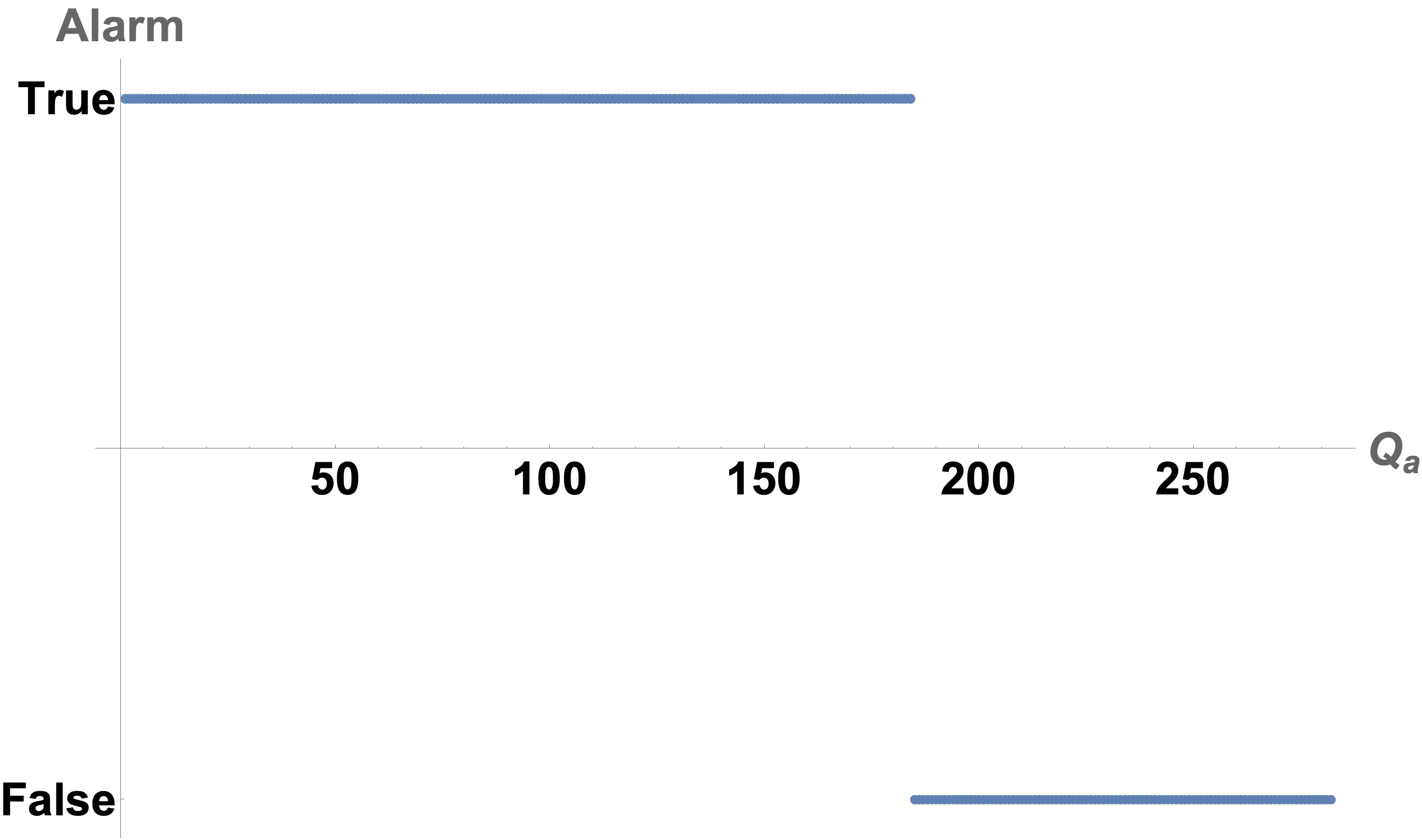}
  \caption{Logical trace for the misalignment alarm when the evaluation
  sketch in Table~\ref{tbl:grading-llms} is transformed to make all the
  LLMs satisfy the safety specification. All three are being compared
  simultaneously at any fixed \qa so the single classifier axiom defines
  a cuboid in the 3-dimensional space for each label.}
  \label{fig:logical-trace-all-correct}
\end{figure}

One can play around with the single example in Table~\ref{tbl:grading-llms} to create alternative or spoofed
test summaries where the classifiers all fail, all are correct, etc. One transformation is flipping
whatever label a classifier produces. This requires no knowledge of the true label. The other two
do require it and would flip labels given the chosen classifier and true label.
A logical trace of the misalignment alarm when all three LLMs are made to satisfy the
safety specification is shown in Figure~\ref{fig:logical-trace-all-correct}. For
this example, the misalignment alarm is not triggered since there exist \qa
values where all classifiers have evaluations that satisfy the safety specification.

It should be clear that a logic of unsupervised evaluation cannot provide
the context in which it is being used safely. How big the test should be
and how much disagreement between classifiers should be considered enough
to alarm must be established in the application context of the alarm.
The safety specification we used as a running example throughout this paper
was arbitrary and could be made harder or relaxed.

\subsection{Detecting spoofed inputs to the evaluation model}

There are various reasons to call the algebraic relations obtained
from the generating set for an ensemble \emph{axiomatic}. As already
noted, they are universal and apply to all binary evaluations. There
are no free parameters in these expressions to learn or train.

One very practical reason to consider them axioms is that their violation
would immediately tell us that something is wrong. We have already
encountered one use of this verification role for the axioms - verify
that the group evaluations calculated by an evaluating algorithm lie
in the logically consistent set. Another use of this verification is
the detection of spoofed test summaries.

We already discussed global transformations of the label decisions
that are guaranteed to be part of the generating set and are thus
undetectable. But other transformations could occur, whether malicious
or not. Can we detect them? This connects the work here further
with notions from error detection and correction. We only remark
briefly on it.

One way to create a spoofed test summary is to use a different
generating set for the observed counts of agreements and disagreements.
For example, we could randomly pick positive integers for the $2^N$
patterns represented by the generating set,
\begin{align}
    \raiaj{i}{j} &= x \\
    \raibj{i}{j} &= y \\
    \rbiaj{i}{j} &= w \\
    \rbibj{i}{j} &= v, 
\end{align}
and $x + y + w + v = Q.$ We can view the generating sets as maps
from unobservable statistics to observable ones. This spoofed generating
set looks quite different from the one associated with a binary evaluation.
Therefore, there is no guarantee that a spoofed test summary would actually
be possible during \emph{any} binary evaluation. This turns detection into
a geometrical problem. Does the generating set with the spoofed summary
have an empty variety - there are no points in \rspace that satisfy the
relations?

\subsection{Exact evaluation for error-independent classifiers}

The generating set for an ensemble of size $N=3$ error-independent
classifiers has the form,
\begin{flalign}
  \freqthree{a_i}{a_j}{a_k}  &=  \prva \prs{i}{a} \prs{j}{a} \prs{k}{a}& + \; \; & \prvb (1 - \prs{i}{b}) (1 - \prs{j}{b}) (1 - \prs{k}{b})\\
  \freqthree{a_i}{a_j}{b_k}  &=  \prva  \prs{i}{a} \prs{j}{a} (1 - \prs{k}{a}) & + \; \; & \prvb (1 - \prs{i}{b}) (1 - \prs{j}{b}) \prs{k}{b}\\
  \freqthree{a_i}{b_j}{a_k}  &=  \prva  \prs{i}{a}  (1 - \prs{j}{a})  \prs{k}{a} & + \; \; & \prvb (1 - \prs{i}{b}) \prs{j}{b} (1 - \prs{k}{b})\\
  \freqthree{b_i}{a_j}{a_k}  &=  \prva  (1 - \prs{i}{a}) \prs{j}{a} \prs{k}{a} & + \; \; & \prvb \prs{i}{b} (1 - \prs{j}{b}) (1 - \prs{k}{b})\\
  \freqthree{b_i}{b_j}{a_k}  &=  \prva  (1 - \prs{i}{a}) (1 - \prs{j}{a}) \prs{k}{a} & + \;\; & \prvb \prs{i}{b} \prs{j}{b} (1 - \prs{k}{b})\\
  \freqthree{b_i}{a_j}{b_k}  &=  \prva  (1 - \prs{i}{a})  \, \prs{j}{a} \, (1 - \prs{k}{a}) & + \;\; & \prvb \prs{i}{b} (1 - \prs{j}{b}) \prs{k}{b}\\
  \freqthree{a_i}{b_j}{b_k}  &= \prva  \prs{i}{a}  (1 - \prs{j}{a})  (1 - \prs{k}{a}) & + \;\; & \prvb (1 - \prs{i}{b}) \prs{j}{b} \prs{k}{b}\\
  \freqthree{b_i}{b_j}{b_k}  &= \prva  (1 - \prs{i}{a})  (1 - \prs{j}{a}) (1 - \prs{k}{a}) & + \;\; & \prvb \prs{i}{b} \prs{j}{b} \prs{k}{b}.
\end{flalign}
This generating set has an algebraic variety (set of points that satisfy the polynomials) that consists of two
points. One is the true performance of the ensemble. The second one is related by the transformations,
\begin{align}
    \pa & \rightarrow (1 - \pa) \\
    \pia{i} & \rightarrow (1 - \pia{i}) \\
    \pib{i} & \rightarrow (1 - \pib{i}).
\end{align}
These solutions are easily obtained in any software package that contains implementations of
Buchberger's algorithm. For example, in the \emph{Wolfram} language, the solution can
be obtained with the built-in function \texttt{Solve} in seconds.

This solution is hardly known in the ML/AI literature. It is an algorithm that uses
the decisions of the ensemble to evaluate itself. As such, it is the \emph{evaluation}
version of the well-known \emph{decision} algorithm for ensembles - majority voting.
Indeed, it is better. Majority voting is known, by Condorcet's theorem, to minimize
decision errors if all classifiers are better than 50\% at making their decisions.
If you knew that \prva was either greater or less than 50\%, you would be able
to perfectly evaluate the ensemble. Table~\ref{tbl:prevalence-comparison} compares
the estimates for label \lbla prevalence from majority voting with that from
this exact solution. This direct comparison makes clear that the exact solution
for error independent classifiers is \emph{not} equivalent to majority voting
evaluations.

\begin{table}
  \centering
  \begin{tabular}{lc}\toprule
    Evaluator     & \pa  \\
    \toprule
    Majority Voting & \mvprev  \\
    \midrule
    Fully inferential  &  \aeprev  \\
    \bottomrule
  \end{tabular}
  \caption{Algebraic evaluation formulas for the prevalence of label
  \lbla for two different evaluators, majority voting and the 
  algebraic solution using the axioms presented here up to $N=3.$
  The formula for the algebraic solution via the axioms has the
  interesting property that it can return irrational values.
  These can serve as alarms that the assumption of the formula,
  error independence, does not hold. The quantities $\Delta_{i,j}$
  are 2nd order moments of the observable responses:
  $f_{b_i,b_j} - f_{b_i} f_{b_j}$ or $f_{a_i,a_j} - f_{a_i} f_{a_j}.$
  In binary classification these two expressions are equivalent.}
  \label{tbl:prevalence-comparison}
\end{table}

This exact solution is notable for another reason related to AI safety - it can
signal the failure of its own assumptions. No probabilistic model assuming
error independence can do this. The generating set for an ensemble of size $N=3$
becomes an evaluating algorithm under the assumptions of error independence between
the classifiers. But this algorithm has no free parameters to train or adjust.
As such, it can return estimates that are clearly wrong. For example, irrational
or complex numbers for any of the sample statistics in \pspace.

Platanios et al \cite{Platanios2014, Platanios2016} derived an expression for
percentage correct answers of independent classifiers for any number of labels.
The error rate of classifier $i$ is given by,
\begin{equation}
    e_i = \frac{c \pm (1 - 2 a_{j,k})}{\pm 2 (1 - 2 a_{j,k})},
\end{equation}
where the $a_{i,j}$ like terms are the agreement rates between
pairs of classifiers.
The difficulty arises in the expression `c' - it contains an unresolved square
root,
\begin{equation}
   c = \sqrt{(1 - 2 a_{1,2})(1 - 2 a_{1,3})(1 - 2 a_{2,3})}.
\end{equation}
By construction, for a finite test, the agreement rates between any set of classifiers
would be an integer ratio. So this term must resolve when the independence condition
applies because that is how this `c' term was derived. And it must resolve for \emph{any}
set of independent classifiers. That is an extraordinary coincidence any of us would be
hard pressed to accept. But that square root should have alerted the humans considering
this theory that something was wrong with it.

The error in the derivation of the Platanios independent solution arose when they
assumed that percentage correct for a pair could be written as the products of
their percentage correct. This is not true in general when we write their
expression with label statistics, not just agreement statistics -
\begin{align}
    e_{i,j}& = e_i e_j \\
    ( \prva e_{i,a} e_{j,a} + \prvb e_{i,b} e_{j,b} )& \neq (\prva e_{i,a} + \prvb e_{i,b}) (\prva e_{j,a} + \prvb e_{j,b}).
\end{align}
Platanios et al confused stream error rates with label error rates.

\subsection{Formalisms for multi-label classification}

The formalism presented here for binary classification, denoted by $R=2$,
can be readily extended to more labels. Binary classification has
the property that there is only one way to be wrong. But with
$R>2$ classifications, there are more ways to be wrong than right.
We could have presented all the formalism for binary classification
in terms of inacurracies rather than accuracies. For three or more
labels the generating sets look more symmetric if we do all the
computations in terms of being wrong about a label. For
example, in $R=3$ evaluations we would use,
\begin{equation}
    \phat_{a_i,a} = 1 - \phat_{b_i,a} - \phat_{c_i,a}
\end{equation}
for the percentage of times \lbla questions where answered
correctly - one minus the percentage of times it was wrong
by saying it was \lblb and $\mathcal{C}.$

Binary classification also makes it possible to talk about
only one error correlation per label. In general we have
to consider many more correlations. Take the case of pair
correlations. Its general form can be written as,
\begin{equation}
    \Gamma_{\ell_i, \ell_j}^{\ell_\text{true}}.
\end{equation}
So we have to consider tensors of correlation statistics.

Any measurement can be digitized to a finite number of
ranges. If one had the formalism for evaluation logic of
tests with $R$ responses, the verification formalism
presented here could be used to make sure agents using
or producing them are working correctly. Evaluation models
of other evaluations are possible so as to simplify them
and thereby gain an easy way to verify them.

\end{document}